\documentclass[runningheads]{llncs}
\usepackage[T1]{fontenc}
\usepackage{amsmath,amsfonts}
\usepackage{algorithmic}
\usepackage{algorithm}
\usepackage{array}
\usepackage[caption=false,font=normalsize,labelfont=sf,textfont=sf]{subfig}
\usepackage{textcomp}
\usepackage{stfloats}
\usepackage{url}
\usepackage{verbatim}
\usepackage{graphicx}
\usepackage{cite}
\usepackage{amssymb}
\usepackage{textcomp}
\usepackage{xcolor}
\usepackage{xspace}
\newcommand{\algoName}{Nodule\xspace}  

\begin{document}

\title{Unsupervised Cognition}

%
\author{Alfredo Ibias\inst{1}\orcidID{0000-0002-3122-4272} \and
Hector Antona\inst{1}\orcidID{0009-0005-8706-1301} \and
Guillem Ramirez-Miranda\inst{1}\orcidID{0000-0003-2741-3705} \and
Enric Guinovart\inst{1} \and
Eduard Alarcon\inst{2}
}
\authorrunning{A. Ibias et al.}
%
\institute{Avatar Cognition, Barcelona, Spain
\email{\{alfredo, hector, guillem, enric\}@avatarcognition.com} \and
Universitat Politècnica de Catalunya - BarcelonaTech, Barcelona, Spain
\email{eduard.alarcon@upc.edu}}
%


\maketitle

\begin{abstract}
Unsupervised learning methods have a soft inspiration in cognition models. To this day, the most successful unsupervised learning methods revolve around clustering samples in a mathematical space. In this paper we propose a primitive-based, unsupervised learning approach for decision-making inspired by a novel cognition framework. This representation-centric approach models the input space constructively as a distributed hierarchical structure in an input-agnostic way. We compared our approach with both current state-of-the-art unsupervised learning classification, with current state-of-the-art small and incomplete datasets classification, and with current state-of-the-art cancer type classification. We show how our proposal outperforms previous state-of-the-art. We also evaluate some cognition-like properties of our proposal where it not only outperforms the compared algorithms (even supervised learning ones), but it also shows a different, more cognition-like, behaviour.
\keywords{Unsupervised learning \and Incremental learning \and Cognitive systems \and Explainable AI \and Computational intelligence.}
\end{abstract}



\section{Introduction}\label{sec:intro}
Unsupervised learning is a huge field focused on extracting patterns from data without knowing the actual classes present in it. Due to this particularity, the field is full of methods that cluster data based on its mathematical representation. This hampers their applicability to data whose mathematical relationships do not directly correlate with its cognitive relationships, relationships that cognitive agents (like humans) find between the data. For example, the MNIST dataset has a clear cognitive relationship between its different samples: the number they represent. However, when transformed into numerical values for clustering, their relationships fade out in favour of relationships between their encodings, that do not necessarily correspond with the cognitive ones.

In this field, there are multiple algorithms that focus on the unsupervised classification problem. However, due to their soft inspiration in cognition models, most of them address the problem from an optimisation perspective. This approach requires building a mapping between any input and a valid output (ideally, the best output), and thus it is dividing the input space into subspaces. In that regard, the representations become spatial, in the sense that the classes are represented by subspaces of an infinite space, independently of the similarity between the inputs that fall in that subspace.

In contrast, novel theories about how the brain works propose that the brain models the world in a constructive way, that is, it generates constructive representations of the world~\cite{hb04, lcgh15, yhp20}. A constructive representation would be an abstraction or archetype of a class, in the sense that, it would be a representation to which any (or at least some) elements of the class are similar to. This implies that, to assign a class to an input, it has to be similar enough to one of the already learned representations and, if it is not similar enough to any of them, it can not be classified. Mathematically speaking, the difference between both approaches is that the first, traditional one focuses on splitting a representation space, and the second, novel one focuses on building a set of representations.

To empirically evaluate this new approach, we need to develop a new unsupervised learning algorithm. To develop such algorithm, we decided to follow a novel cognition framework (Synthetic Cognition)~\cite{irga24} that is based in the previously mentioned theories of the brain. This framework presents the \emph{Self-Projecting Persistence Principle} (SPPP), that defines how latent information is present in reality, how it persist in time, and how it projects itself to the world through manifestations. As such, this framework presents the basic cognition task as building abstractions based on manifestations captured from latent information, with the goal that such abstractions approximate the original latent information. To fulfil such task, the framework proposes to process manifestations into constructive representations through a primitive-based processing. Finally, this primitive-based processing is scaled to build a whole Cognitive Architecture.

Our aim in this paper is not to develop the full Cognitive Architecture, but just the Perception~\cite{llr17} part. This Perception should recognise inputs without needing a label, and hence it is equivalent to an unsupervised learning algorithm. The previously mentioned cognition framework~\cite{irga24} defines some requirements for developing such an algorithm: it has to be input-agnostic, primitive-based, scalable, and representation-centric. With these requirements, in this paper we propose a novel unsupervised learning algorithm for decision-making we call \emph{\algoName}. We expect this algorithm to be a building block towards a full cognition algorithm based on the previously mentioned cognition framework. 

In this paper we present the fundamentals of the \algoName as well as one of its modulators: the \emph{Spatial Attention} modulator. This modulator will auto-regulate the spatial discriminability of the algorithm. Additionally, we developed our proposal to be transparent and explainable, as it is desirable that any solution can describe its representations and explain its decisions. Finally, a perk of focusing on generating constructive representations is that our algorithm is able to state if a new input does not correspond to any previously seen pattern, that is, it can say ``I do not know''.

We compared our proposal to the main unsupervised classification algorithms: K-Means for tabular data and Invariant Information Clustering (IIC)~\cite{jvh19} for image data. We compared it with different configurations of K-Means and IIC and for four different static datasets (two tabular and two image datasets), for an unsupervised classification task. The results show a clear advantage of our proposal, being able to deal with both tabular and image data with a decent performance. We also performed a comparison over small and incomplete datasets in which we beat current state-of-the-art, and another comparison with a medical dataset in which we beat the current state-of-the-art for classifying cancer types. Finally, we performed some experiments to evaluate cognition-like properties. In this case we compared our proposal to not only K-Means and IIC, but also the K-NN clustering supervised method. The comparison consisted on recognising MNIST digits even when removing random pixels. In that experiment there is a clear advantage of our proposal over the compared algorithms that shows how building constructive representations produces a different behaviour, and thus has the potential to have cognition-like properties. Given these results we conclude that our proposal is a state-of-the-art disruptive unsupervised learning algorithm for decision-making, with different, more promising properties than traditional algorithms.

The rest of the paper is organised in a related work summary at Section~\ref{sec:rel}, a proposal description in Section~\ref{sec:prop}, an empirical evaluation at Section~\ref{sec:exp}, a discussion in Section~\ref{sec:disc},
and a summary of the conclusions and future work at Section~\ref{sec:conc}.

\section{Related Work}\label{sec:rel}
There are multiple algorithms for unsupervised learning developed along the years, from generic clustering algorithms like K-Means~\cite{Lloyd82}, to more specific, usually Artificial Neural Network-based, algorithms that deal with only one task. In this second category we can find algorithms that deal with topics as unrelated as representation learning~\cite{wlgs20, stsc24}, video segmentation~\cite{asr21}, speech recognition~\cite{bhca21} or community detection~\cite{gcoalzl24}. However, none of them try to build constructive representations, but instead they divide a mathematical representation of the input space into clusters that represent the different classes present in the input space.

Among these clustering algorithms, there are few that stand out, specially for the task of unsupervised classification. One of them is K-Means due to its performance clustering tabular data. This algorithm clusters the samples based on their closeness in the mathematical space of their encodings. Another one is Invariant Information Clustering (IIC)~\cite{jvh19} due to its performance clustering images. This algorithm takes an image, transforms it with a given transform, and then run both of them over two Artificial Neural Networks with the goal of learning what is common between them. To that effect, it aims to maximise the mutual information between encoded variables, what makes representations of paired samples the same, but not through minimising representation distance like it is done in K-Means. In any case, both algorithms stand out due to their performance in their respective domains, but none of them is able to obtain good accuracy across domains. Thus, we will use them as baseline for comparison purposes, even though they cannot be applied in all cases.

Regarding approaches relevant to mention to provide context for our contribution, it is important to highlight the following ones. On the one hand, the way our primitive-based architecture constructs representations shows similarities, in other application domains, with the so-called Agglomerative Hierarchical Clustering algorithms, which were originally discussed in~\cite{mullner11} and recently being revisited in a structured manner in~\cite{oo24}. These methods are based upon assigning pair-wise dissimilarities among data points to generate a data structure in the form of a tree-like dendrogram. Similarities to our approach consist of a growing policy based upon similarities of vector contents, despite the application not being for representations nor cognitive systems. On the other hand, the work introduced in this paper constitutes a step towards deriving structured representations of world models, akin to the concepts introduced in~\cite{hs18} and elaborated and reinforced recently in~\cite{zplp24}.

In addition, our proposal of constructive abstracted representations that constitute the basis for finding distances to input sensorial data has conceptual similarities to the body of work in the field of neuro-symbolic representations, as surveyed in~\cite{bgbbdh+21} and more recently in~\cite{chtwsr24}. This top-down representation matched with lower-level architectural implementations (which naturally map to underlying in-memory computing concepts) are also discussed in hyperdimensional computing models or vector symbolic architectures, as discussed in~\cite{kslcbsr21,kror23,pazingi22} culminating in a recent patent~\cite{hsr24}.

Finally, regarding brain-inspired methods that try to model the input space, the only research we are aware of is the Hierarchical Temporal Memory~\cite{cah17} (HTM) and SyncMap~\cite{va21}, although they are algorithms suited for learning sequences instead of static data, and HTM is not unsupervised. Thus, as far as we are aware, ours is the first proposal of a brain-inspired, primitive-based, unsupervised learning algorithm for modelling static data.

\section{Unsupervised Cognition}\label{sec:prop}
Our proposal, based on the novel cognition framework presented at~\cite{irga24}, is composed by: an Embodiment, a \algoName, and a Spatial Attention modulator. The goal of the Embodiment is to transform the input space into Sparse Distributed Representations (SDRs), the goal of the \algoName is to process those SDRs and model the input space generating constructive representations, and the goal of the Spatial Attention modulator is to auto-regulate the \algoName.

\subsection{The Embodiment}\label{subsec:emb}
The goal of our unsupervised learning algorithm is to model the input space generating constructive representations. To do so, it requires a representation-oriented universal data structure. Recent research has shown that such data structure is Sparse Distributed Representations~\cite{ah16,cah17} (SDRs), which allows for input-agnostic representations of inputs independently of their data type. This has been proven to be the actual way in which the brain processes its inputs~\cite{foldiak03,cah17}. Thus, our algorithm will work only with SDRs.

\begin{figure}[!t]
    \centering
    \includegraphics[width=0.4\columnwidth]{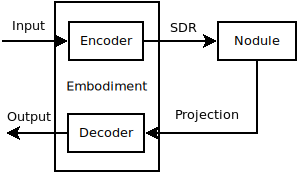}
    \caption{\small Global Schema}
    \label{fig:glob}
\end{figure}

To translate inputs to SDRs we need an encoder architecture. To interpret the SDRs the \algoName generates we need a decoder architecture too. Both architectures conform the \emph{Embodiment} of the \algoName, as displayed in Figure~\ref{fig:glob}. In our case, as in our experiments we only explore tabular and image datasets, we only present four kinds of encoder decoder pairs: one for float point numbers, one for categorical data, one for grey images, and one for colour images. All these encoders are lossless and thus allow us to recover the encoded data.

The grey images translation to SDR is straightforward: a grey image's SDR is a flattened version of the image (in which each pixel is a dimension) with the values normalised to be between $0$ and $1$. In the case of colour images, we perform the same transformation to each one of the RGB channels and we concatenate their SDRs to form the image SDR.

For floating point numbers their translation to SDR is a bit more nuanced. We take the input space of the number (that is, the possible values it can take) and divide it into bins. Those bins will be the dimensions of the SDR. Then, each number will fall into one of those bins. However, in order to allow some overlap between numbers (fundamental for finding similarities using \algoName), we also activate as many bins around the number bin as another parameter we call \emph{bin overlap}. With this, the SDR is a long list of zeros and some ones around the bin where the number falls. By default, we use an overlap of $10\%$ of the number of bins, that by default is set to $100$ bins. In the case of categorical data we create one bin per category and set the overlap to $0\%$, following a one-hot encoding.

Having these representations, we define the SDR representation of a tabular input as a concatenation of the SDR representations of each entry, adjusting the indices to put one entry representation after another. Using this same methodology, we can compose multiple input types into one SDR. Although this is not the goal of this paper, these compositions could potentially help our algorithm to deal with datasets in which the input has many different types (for example, keyboard keys and video, like the recently released MineRL dataset~\cite{ghtwcvs19}), although proving that would be matter of future work.

Here it is important to remark that, due to the transformation of any input and any output into SDRs, the algorithm always deals with the same data structures, and thus it is optimised to learn them independently of what they represent. Moreover, this makes our algorithm input-agnostic, as any kind of data is potentially transformable into an SDR.

\subsection{The \algoName}\label{subsec:mem}
Once we have an encoder-decoder pair, we need to process the SDRs they generate. To that end, and following the requirements outlined at Section~\ref{sec:intro}, we need a primitive-based processing. We defined a basic primitive we called \emph{Footprint}, and we defined its scalability grouping multiple Footprints into a set we called \emph{Cell}, and grouping multiple Cells into a hierarchy we called \emph{Nodule}. Footprint and Cell are terminology coming from~\cite{irga24}, while Nodule is a more aseptic name we give to the \emph{Cluster} defined at~\cite{irga24}.

\subsubsection{The Footprint}
\begin{figure}[!t]
    \centering
    \includegraphics[width=0.2\columnwidth]{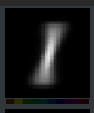}
    \includegraphics[width=0.65\columnwidth]{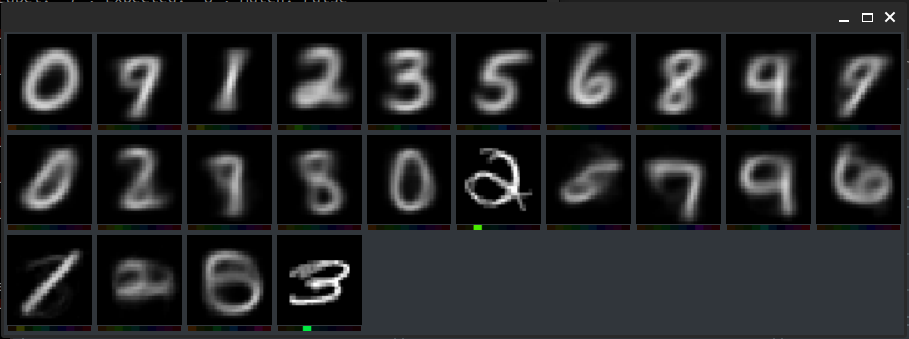}
    \caption{(left) An example of Footprint: the combination of the 1's of the MNIST dataset.\\(right) An example of Cell: the Footprints of the 60,000 samples of the MNIST dataset.}
    \label{fig:FP}
\end{figure}

Our most basic processing unit, that is, our primitive, is the Footprint. A Footprint is an internal representation with two basic functions. A Footprint contains a data SDR, and can contain (for evaluation purposes only) a metadata SDR (i.e. an SDR representing a label). A Footprint also has an \emph{updating function} and an \emph{activation function}.
The updating function modifies the data and metadata SDRs by computing the running average between the Footprints SDRs and the external SDR, while the activation function computes the average between the Footprint SDRs and the external SDR in what we call the \emph{Archetype} of such Footprint. An example of a Footprint is displayed at Figure~\ref{fig:FP} left.



The Archetype is the output of a Footprint after processing an input, and it is derived from the SPPP~\cite{irga24}. The Archetype aims to be an abstracted version of the input, and the manifestation (and thus self-projection) of the Footprint. It is key to build hierarchies with Footprints.


\subsubsection{The Cell}
The main limitation of Footprints is that they can only store one representation. Thus, to be able to have multiple representations, we needed to create multiple Footprints. To organise them, we grouped them inside a structure called Cell, that coordinates the Footprints to avoid redundancies. Whenever a Cell receives an input, it has to decide which Footprint will process it to produce the outputs, and it does so trough similarity: the most similar Footprint is the one that is activated, and thus the one that will process the input, and whose Archetype will be the Cell's Archetype. Thus, a Cell contains a set of Footprints and a threshold common to all its Footprints. This threshold is used to decide whether an input is similar to an existing Footprint or not, and will be deeply defined in the next section. An example of a Cell is displayed at Figure~\ref{fig:FP} right.


To decide if a new input is considered similar to an existing Footprint, we use a Similarity Function to get an score of the similarity of both SDRs, and if that score is over the Cell's threshold, then both the input and the Footprint are considered similar to each other. When there are more than one Footprint with a similarity over the Cell's threshold, we consider similar only the one with higher similarity. This similar Footprint is then the \emph{active} Footprint.

To compute the previously mentioned similarity score between SDRs we need a \emph{Similarity Function} that compares SDRs. This Similarity Function should take two SDRs and return a similarity value stating how similar we consider them to be. The specific Similarity Function we developed for this paper is a variation of the euclidean distance, but we also tested using the euclidean distance between vectors and the differences in results are minimal. An important remark here is that the Similarity Function should use only the data part of both SDRs to compute the similarity in order for our method to be fully unsupervised, leaving the metadata part outside of any decision making.

\subsubsection{The Nodule}
\begin{figure}[!t]
    \centering
    \includegraphics[width=0.5\columnwidth]{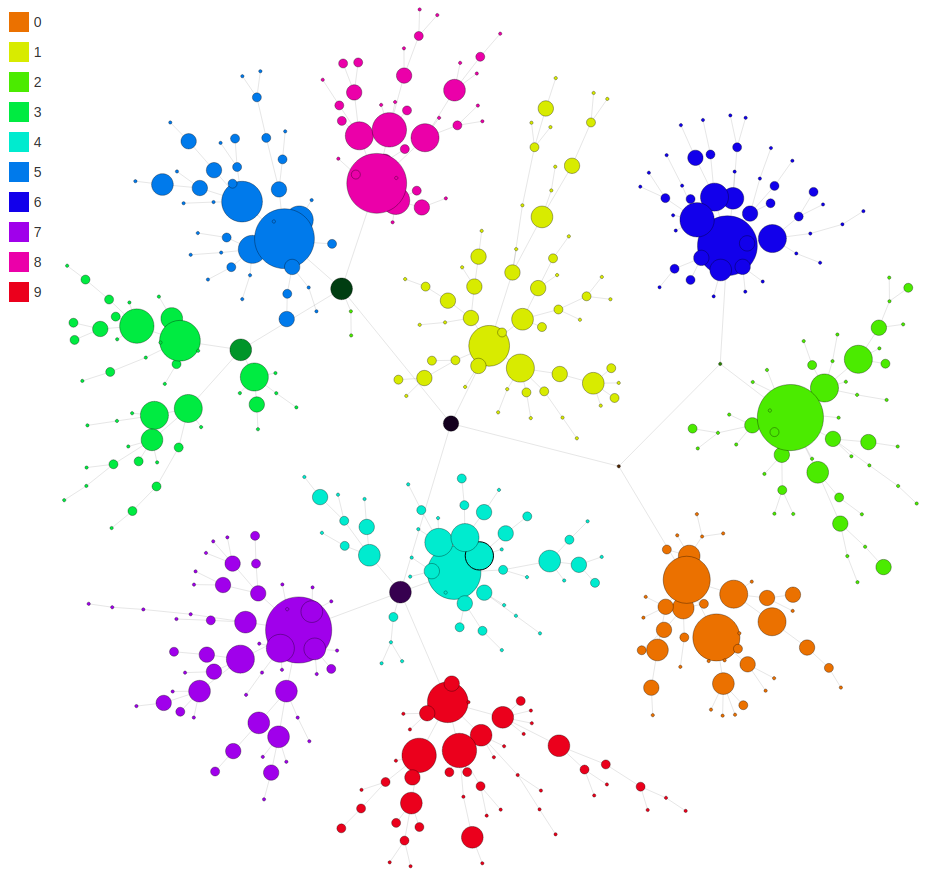}
    \caption{An example of Nodule: the Cells of the hierarchy present in the 60,000 samples of the MNIST dataset. The Seed Cell is the central black circle.}
    \label{fig:Nodule}
\end{figure}

The core of our algorithm revolves around properly building and organising Footprints. To organise them, they are grouped inside Cells, but a Cell only allows us to have representations at the same level of abstraction. To have multiple levels of abstractions, we need to have multiple Cells organised in a tree-like hierarchical structure, and that is what we call Nodule. An example of Nodule is displayed at Figure~\ref{fig:Nodule}.

At the beginning of training there is only one Cell in the Nodule (which we call Seed Cell), that starts with no Footprints. When new inputs are processed during training, new Footprints are generated and the Seed Cell grows. And when an input is considered similar to an existing Footprint, then the need for a hierarchy arises. In such a case, a new Cell is created as a child of the Seed Cell, and it is associated with that Footprint. Thus, a Cell can have as many children as Footprints, and each child Cell has an associated parent Footprint.

With this organisation, a Nodule's goal is to organise Cells in a subset hierarchy, in which the Seed Cell contains the Footprints representing more inputs, and the leaf Cells contain Footprints representing only one input. The idea is that any child Cell will subdivide the subset of inputs represented by its parent Footprint. To that end, is fundamental the fact that each Cell has its own similarity threshold, what allows for a better discrimination policy as we will see in following sections.

\subsubsection{Processing an Input}
\begin{figure}[!t]
    \centering
    \includegraphics[width=\columnwidth]{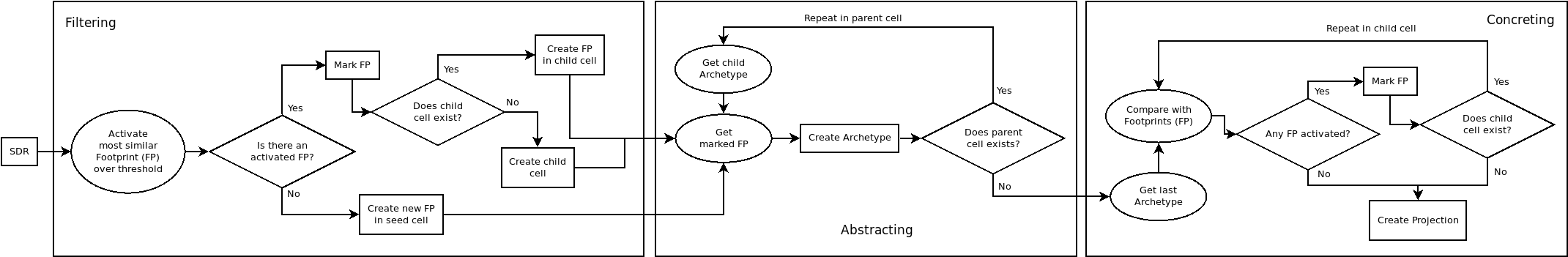}
    \caption{\small The \algoName's Training Schema}
    \label{fig:algo}
\end{figure}

Now let us show how a new input is processed by the \algoName. To follow this description, a general schema of this algorithm is displayed at Figure~\ref{fig:algo}. As we can see in the schema, our input processing method has two phases. The first phase is a filtering one, in which we look for the Footprint most similar to the input. The second phase is an abstracting one, in which we go up the Nodule generating an abstraction of the input using the Cell's Archetype. Is in this last phase where the Footprint update happens, what is key for our generalisation potential.

We start the filtering phase (left side of Figure~\ref{fig:algo}) with an SDR that is a new input (this SDR contains both a data part and a metadata part). This input is then compared to all the Footprints present in the Nodule, and we select, from the Footprints that surpass their Cell's similarity threshold, the Footprint that gets the highest similarity. If a Footprint is selected, then we check if it has a child Cell, and if it has it, we create a new Footprint copy of the input in such Cell. If the Footprint does not have a child Cell, then one is created that contains two new Footprints: one is a copy of the input, the other is a copy of the parent Footprint. If no Footprint is selected, then a new Footprint copy of the input will be created in the Seed Cell and will be selected. If we are in evaluation mode, no Cell or Footprint are created in this phase, but the selected Footprint will be the result of our algorithm, what we call the \emph{Projection} of the Nodule.

In the abstracting phase (centre side of Figure~\ref{fig:algo}), the selected Footprint is activated and executes its updating and activation functions in that order. The Archetype produced by the activation function is then passed up as input to its parent Cell, where the parent Footprint is activated and executes its updating and activation functions with its child Cell's Archetype. This way a Footprint update is performed with an aggregation of the input and the active child Footprints. This process is repeated until it has been done in the Seed Cell, and the Archetype generated by the Seed Cell is considered the Archetype of the Nodule. If we are in evaluation mode, no Footprint is updated in this phase.


The output of the whole method is the Projection of the Nodule, that includes a data SDR with the representation most similar to the input and a metadata SDR with the additional information about the data SDR provided (like a label). This Projection is later processed by the decoder of the Embodiment to retrieve the final label of the input. A resume of the global algorithm is displayed at Figure~\ref{fig:glob}. It is important to note that, if we are in evaluation mode, then the update function of the Footprints is not executed to not modify the internal representations, and no new Footprints neither Cells are created. Thus, if there is no Footprint activated in the first phase then there is no Projection and our method returns an ``I do not know''.

Here it is important to remark the relevance of the second phase: it allows generalisation. Without it, our algorithm would be a set of copies of the inputs, deciding the class by the most similar one. That is, it would be a convoluted implementation of the K-Nearest Neighbours algorithm with $K=1$. Adding this phase allows us to build constructive representations, that in turn may be more similar to potential inputs than the already processed ones. Doing the updates in an intelligent way, we have more potential for generalisation. This also allows us to avoid overfitting in subsequent epochs, because the aggregated representation will never be equivalent to an individual input.



\subsection{The Spatial Attention Modulator}\label{subsec:SA}
In the previous Section, a threshold was used to decide if two SDRs were considered similar or not. This threshold can be set arbitrarily, but that would hamper the performance of the \algoName and would generate multiple extra parameters of the model. Thus, a way to automatise the threshold selection was needed, and that is the role of the Spatial Attention Modulator. The whole role of this modulator is to measure the variability of the input space of the Cell and dynamically set a similarity threshold. The algorithm we developed to set such threshold is the average of the mean similarities between the Footprints of the Cell.


The rationale behind using this approach is that any input space has a certain variability, and the right threshold will be that one that sits in the middle of such variability. Such variability is unknown, but the Footprints have captured part of it in the form of aggregations. Thus, each aggregation (that is, each Footprint) represents a class and the average of the similarities between them is the ``captured'' variability.

This actually allows for child Cells to have higher thresholds than their parent Cells, as their input space are limited to those inputs that are similar to their associated parent Footprint. This generates an increase in discrimination power the further down the Nodule a Footprint is. In turn, this develops a distributed hierarchy, in which each Cell processes a different subdomain of the input domain.

\section{Empirical Evaluation}\label{sec:exp}
To evaluate our proposal, we performed four different experiments: a comparison in classification task versus other unsupervised learning algorithms, a comparison in learning efficiency against other methods developed to deal with small and/or incomplete datasets, a state-of-the-art comparison over a medical dataset, and a comparison in cognition-like capabilities versus other clustering algorithms. All the experiments were run in an Ubuntu laptop with an Intel Core i9-13900HX at 2.60GHz with 32 cores, 32Gb of memory, and a NVIDIA GeForce RTX 4060 with 8Gb of VRAM.

\subsection{Experimental Subjects}
Our experimental subjects for these experiments were ten datasets: seven tabular datasets full of numerical values, two image datasets, and a medical data dataset. The datasets and its different properties are presented in Table~\ref{tab:subj}.

\begin{table}[!t]
  \centering
  \caption{\small Characteristics of the experimental subjects}
  \label{tab:subj}
  \resizebox{\columnwidth}{!}{%
  \begin{tabular}{llll}
    \hline
    Name & Type & \# Features & \# Samples \\
    \hline
    Sonar~\cite{sonar, dg17} & Tabular (Numerical) & $60$ & $208$\\
    Ionosphere~\cite{swhb89, dg17} & Tabular (Numerical) & $34$ & $351$\\
    Cancer Type~\cite{wcmsoesss13} & Tabular (Numerical) & $1500$ & $398$\\
    Wisconsin Breast Cancer~\cite{wsm95, dg17} & Tabular (Numerical) & $30$ & $569$\\
    Pima Indians Diabetes~\cite{sedkj88} & Tabular (Numerical) & $8$ & $768$\\
    Air Pressure System Failure~\cite{dg17} & Tabular (Numerical) & $170$ & $60,000$\\
    HIGGS~\cite{bsw14, dg17} & Tabular (Numerical) & $28$ & $100,000$\\
    Accelerometer~\cite{sfss19, dg17} & Tabular (Numerical) & $4$ & $102,000$\\
    MNIST~\cite{lbbh98} & Image (B\&W) & $28\times28$ & $60,000 + 10,000$\\
    CIFAR10~\cite{kh09} & Image (RGB) & $32\times32$ & $50,000 + 10,000$\\
    \hline
  \end{tabular}
  }
\end{table}

We divided some of these datasets into a training set and a test set. For Wisconsin Breast Cancer and Pima Indians Diabetes we split the samples into $70\%$ for the training set and $30\%$ for the test set. In the case of the MNIST and CIFAR10 datasets, they come with $10,000$ samples for test. Thus, we took as training set all the samples from the training dataset and the test set are those $10,000$ test samples. The used Embodiments are the ones described in Section~\ref{subsec:emb}, with an overlap of $10\%$ for Wisconsin Breast Cancer and of $20\%$ for Pima Indians Diabetes due to their respective characteristics.

\subsection{Experiments}
\subsubsection{Learning Curves Experiment}
The first experiment we performed aimed to test how well our proposal deals with a classification task compared to other unsupervised learning algorithms. For tabular data we compared to K-Means with as many centroids as labels, and with the number of centroids that the elbow method~\cite{Thorndike53,ld21} proposes. To evaluate its classification power, each cluster was assigned the label that was most repeated between the training elements of that cluster. In the case of our proposal, the label selected is the one associated to the Projection. When comparing over the image datasets (MNIST and CIFAR10), the Invariant Information Clustering (IIC) algorithm was computed. In this case, the IIC algorithm was setup with the recommended parameters set by the authors for each dataset, and we compared different number of epochs ($1$, $10$, $100$) because we could not try the author recommended number of epochs ($3,200$ for MNIST and $2,000$ for CIFAR10) or any higher number of epochs due to resource constraints.

\begin{figure}[!t]
    \centering
    \hfill
    Wisconsin Breast Cancer
    \hfill
    Pima Indians Diabetes
    \ \ \ \ \ \ \ \ \ \ 
    \ \ \ \ \ \ \ \ \ \ 
    \\
    \includegraphics[width=0.9\columnwidth]{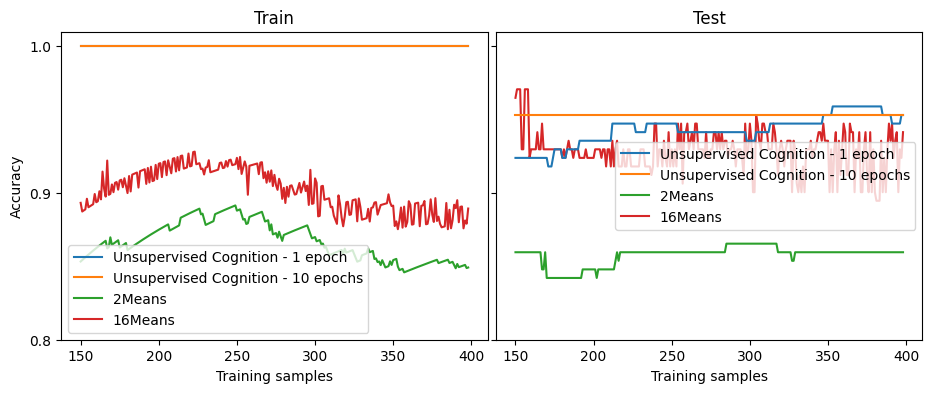}
    \includegraphics[width=0.9\columnwidth]{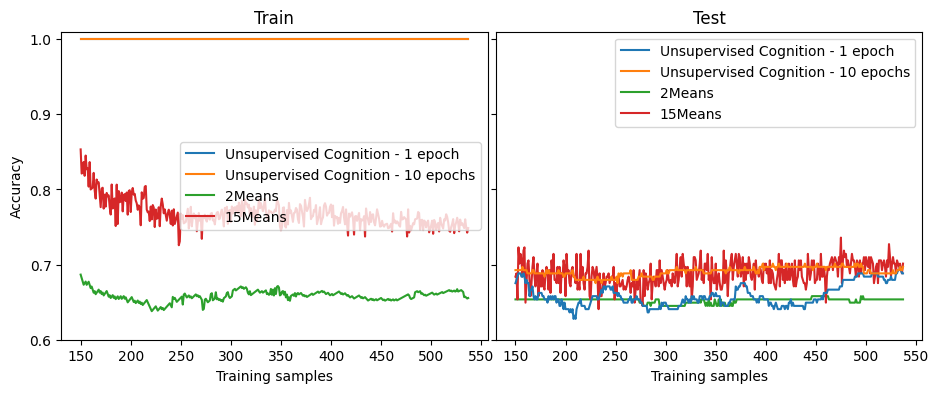}\\
    \hfill
    MNIST
    \hfill
    CIFAR10
    \ \ \ \ \ \ \ \ \ \ 
    \ \ \ \ \ \ \ \ \ \ 
    \ \ \ \ \ \ \ \ \ \ 
    \\
    \includegraphics[width=0.9\columnwidth]{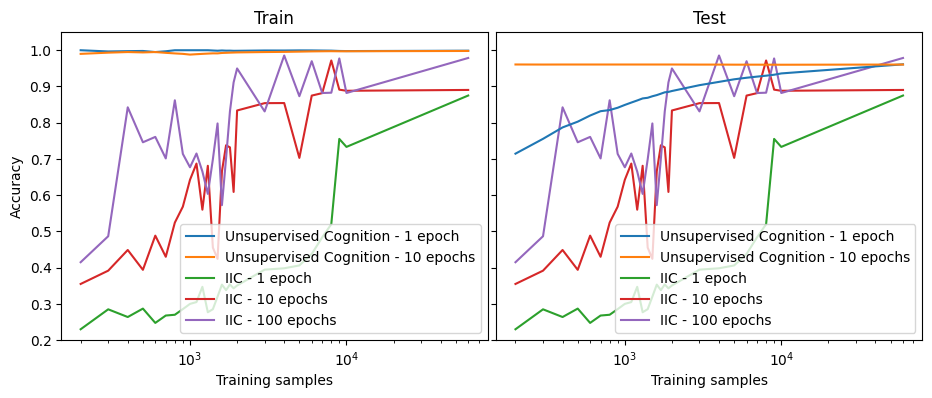}
    \includegraphics[width=0.9\columnwidth]{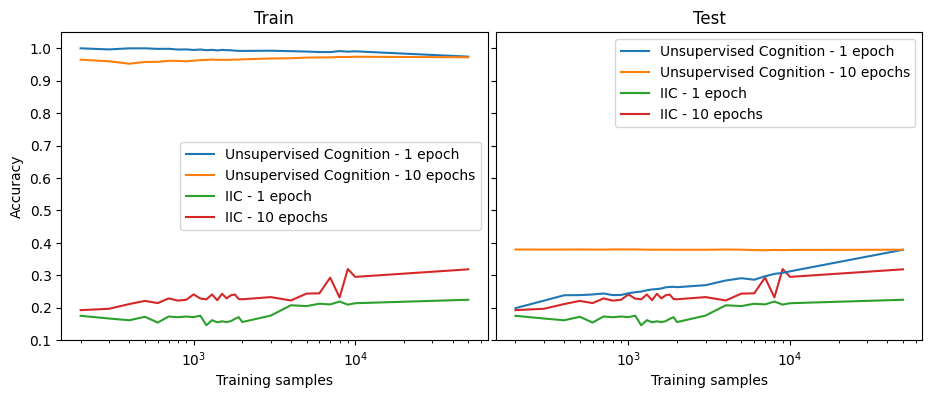}\\
    \caption{\small Learning curves comparison for the different datasets}
    \label{fig:exp1}
\end{figure}

To compare these algorithms, we executed them over the experimental subjects computing the learning curve. That means, we trained the algorithms with the first $150$ samples of the training set, we evaluated them with the samples used to train, and then we tested them over the whole test set. Then we trained them with the first $151$ samples of the training set, evaluated them with the samples used to train, and tested them over the whole test set again, and so on and so forth. We repeated this process, adding $1$ training sample each time, until the whole training set was used for training. We display the resulting learning curves in Figure~\ref{fig:exp1}. Due to the size of the datasets, when computing the image datasets results we executed the experiments for the first $200$ samples, from then on each $100$ samples until the $2,000$ sample, from then on each $1,000$ samples until the $10,000$ sample, and from then on we computed the result with the whole dataset.

The results of this experiment clearly show that our proposal is a better option for unsupervised classification. As we can observe, for tabular data our alternative is on par with K-Means for the Pima Indian Diabetes dataset (loosing by a $1.3\%$) and for the Wisconsin Breast Cancer dataset (winning by a $1.17\%$). When we move to image data, we can observe how our proposal is on par with IIC (loosing by $1.75\%$ for MNIST with $100$ epochs, but winning by $6.05\%$ for CIFAR10 and by $7.07\%$ for MNIST with $10$ epochs).

We want to explicitly remark the fact that our proposal is able to obtain very good accuracies with fewer samples. For contrast, IIC needs around $1,700$ training samples to obtain an stable accuracy over $70\%$ in test MNIST, while our proposal needs less than $200$ samples. Moreover, our proposal does not need multiple epochs to obtain such results: it only goes through the training samples once, although more epochs also improve results (as shown by the ``$10$ epochs'' lines). Finally, if we compare with IIC with only $1$ training epoch, then IIC is not able to overcome our proposal in any scenario, what shows the performance improvement and data efficiency of our approach.

\subsubsection{Small and Incomplete Datasets Experiment}
The second experiment we performed aims to evaluate how well our proposal learns when provided with small and/or incomplete datasets~\cite{ivcgs23}. This experiment focus on the usefulness of our approach in a very common scenario, specially in fields like medical research: when we have a low number of samples, and those samples may not be full. This is an scenario where typical Machine Learning methods can not be applied, either because they cannot work with incomplete samples, or because the low number of samples impede a meaningful learning. Thus, the state-of-the-art in this situation are the widely known Random Forest~\cite{breiman01}, and a recently published method called SANDA~\cite{ivcgs23}.

\begin{figure}[!t]
    \centering
    \includegraphics[width=0.9\columnwidth]{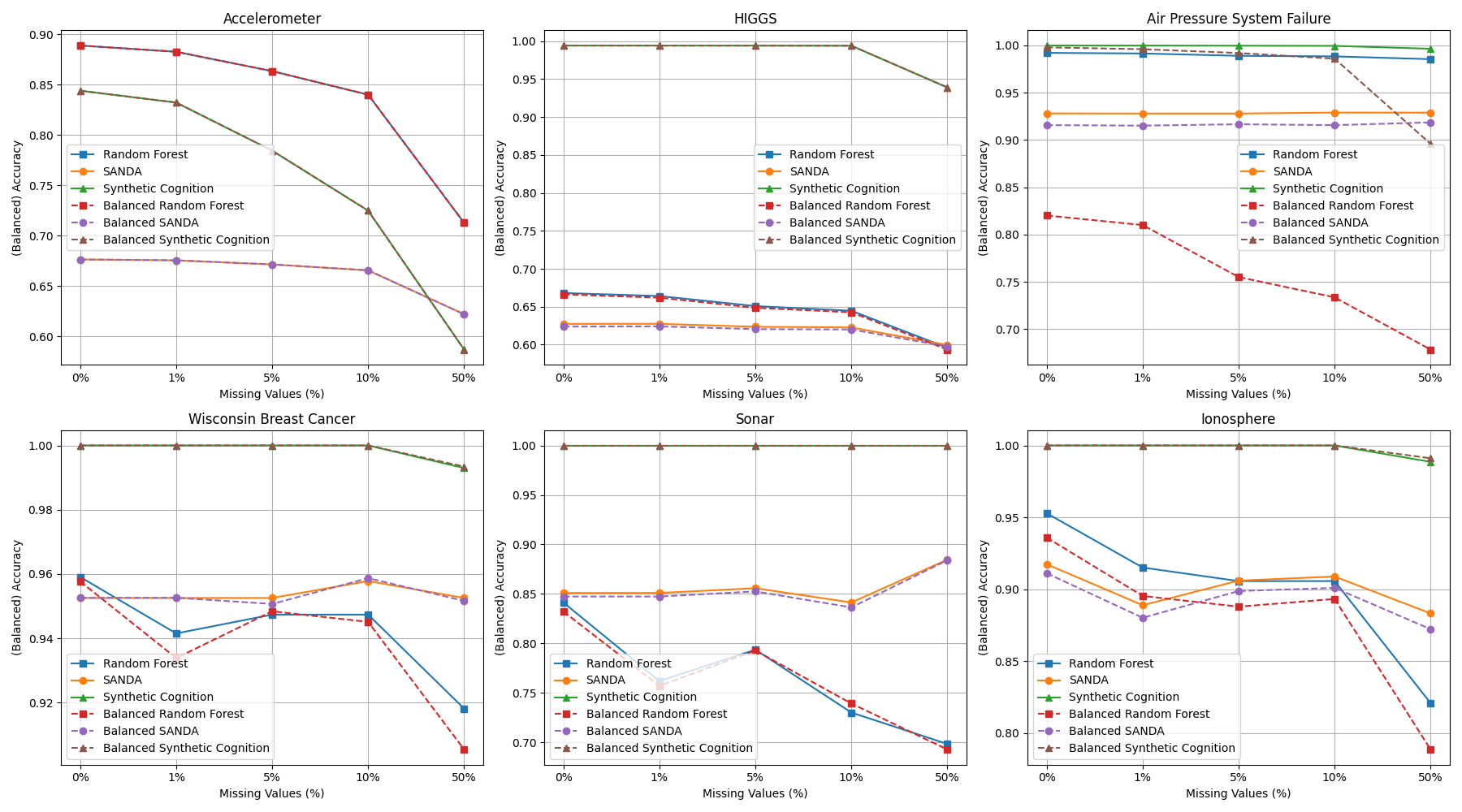}
    \caption{\small Small and incomplete datasets experimental results. Other methods results are from~\cite{ivcgs23}.}
    \label{fig:sanda}
\end{figure}

Hence, in this experiment, we recreated the conditions for the comparison performed in~\cite{ivcgs23}, where six small and medium datasets (Sonar, Ionosphere, Wisconsin Breast Cancer, Air Pressure System Failure, HIGGS, and Accelerometer) were used to simulate an scenario where different levels of data were randomly missing from the samples: $0\%$ (no missing data), $1\%$, $5\%$, $10\%$ and $50\%$ (half of the data is missing). We reproduced the exact same experiment, including the evaluation over the train dataset, and obtained the results displayed in Figure~\ref{fig:sanda}.

As we can observe there, our proposal outperforms both methods, specially when data is missing. Thus, that reveals how well our proposal has been able to learn the dataset, even when data was missing from the samples.

\subsubsection{State-Of-The-Art Experiment}
Now, for our state-of-the-art experiment, we compared our algorithm over the Cancer Type dataset with the state-of-the-art methods evaluated at~\cite{rgdqlyll24}. We performed exactly the same experiment: cancer type classification with five-fold cross-validation. Here being unsupervised was not a requisite, but even with our unsupervised approach we managed to get the results displayed at Table~\ref{tab:sota}, where it is clear how our proposal outperforms the other algorithms (except for the F1 macro measure).

\begin{table}[!t]
  \centering
  \caption{\small Results of multi-class classification between cancer sub-types. Other methods results are from~\cite{rgdqlyll24}}
  \label{tab:sota}
  \begin{tabular}{llll}
    \hline
    Model & Accuracy & F1 weighted & F1 macro \\
    \hline
    GCN~\cite{lmlhlhz22} & $0.73$ & $0.721$ & $0.525$\\
    GAT~\cite{xylzzghy21} & $0.733$ & $0.725$ & $0.552$\\
    MOGONET~\cite{wshtzdh21} & $0.712$ & $0.717$ & $0.614$\\
    MVGNN~\cite{rgdqlyll24} & $0.735$ & $0.725$ & \textbf{0.636}\\
    Unsupervised Cognition & \textbf{0.746} & \textbf{0.737} & $0.513$\\
    \hline
  \end{tabular}
\end{table}

This experiment is crucial to show how our proposal can produce state-of-the-art results in certain scenarios, even when competing against Artificial Neural Network-based models. It also shows how it can be used in real-world scenarios and not only in toy examples like the ones used for the learning curves experiment.

\subsubsection{Cognition-like Capabilities Experiment}
\begin{figure}[!t]
    \centering
    \includegraphics[width=1\columnwidth]{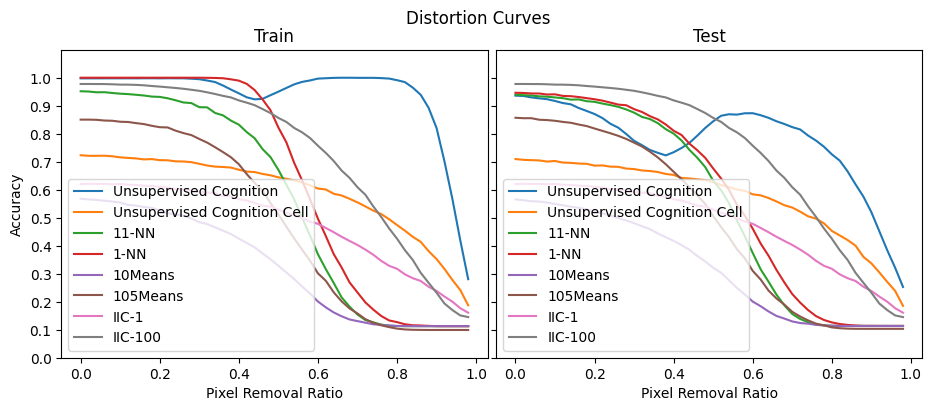}
    \caption{\small Distortion curves comparison for the different clustering algorithms}
    \label{fig:exp2}
\end{figure}

Finally, in our last experiment we wanted to explore the cognition-like capabilities of our proposal, compared to other clustering algorithms, using noise distortion~\cite{zbhrv21}. To that end, inspired by~\cite{admbvrlb23}, we devised an experiment using the MNIST dataset that consists on training the algorithms with the first $10,000$ samples of the training set, and then take the $10,000$ samples from the test set and start taking out pixels. That is, for different percentages (from $0\%$ to $100\%$ with a step of $2\%$), we remove that percentage of pixels (that is, we set them to black) from all the samples of the test set. Then, we evaluate all the algorithms over that test set and compute both the accuracy curve and the area under such curve. We did the same experiment also using the $10,000$ train samples, in order to also evaluate such distortion curve over the already experienced samples. The selected clustering algorithms are: our proposal, our proposal capped to have only $1$ Cell, K-Means with $10$ centroids, K-Means with $105$ centroids, IIC with $1$ epoch, IIC with $100$ epochs, K-NN with $11$ neighbours and K-NN with $1$ neighbour. We display the resulting distortion curves at Figure~\ref{fig:exp2}.

The idea behind this experiment is that, even after removing some pixels from an image, humans are able to recognise numbers. Moreover, if given some specific pixels, and after being told that such pixels represent a number, humans are able to fill in the number. Thus, we understand that recognising and/or reconstructing numbers is a capability of cognitive systems, one that we would desire in any Artificial Intelligence algorithm. Therefore, the goal here is to analyse how well each algorithm is able to recognise and reconstruct numbers from a set of pixels. As the pixels are removed at random, it is expected that after some removal percentage not even humans are able to recognise them, but the more pixels are removed, the better the concept of a number is understood if the number is correctly recognised. Thus, this experiment is expected to set a difference between those algorithms that have an optimisation approach and those that have a modelling one.

As we can observe, our proposal has better distortion curves than any other alternative. In numbers, our proposal obtains an Area Under the Curve (AUC) of $92.60$ over train and of $76.23$ over test. This AUC is way ahead of the next best one, obtained by the IIC with $100$ epochs, that obtains an AUC for both train and test of $72.28$. Moreover, as we can observe in Figure~\ref{fig:exp2}, the other tested methods have a steep descend when there is more than $50-60\%$ of pixels removed, while our proposal keeps getting high accuracy until the very end, when it is almost impossible to recognise the numbers even for a human (around an $80\%$ of pixels removed). This contrast allows us to conclude that the behaviour presented by our proposal is fundamentally different than the behaviour of the alternatives.

We understand that this is a consequence of developing the algorithm around the idea of modelling the input space generating constructive representations.
This effect also implies that our algorithm is finding a different kind of relationships between the samples than the pure numerical or pattern based ones, showing cognition-like properties. Analysing it more in deep, we think that one of the keys of this behaviour is the discriminative hierarchy of patterns. This hierarchy comes from the subdivision of the input space into subspaces through the automatic replication of the Cells, and it allows for a more robust representation of concepts, and thus a better adaptation to noise. Having different levels of representations allows for a better matching between noisy samples and the internal representations, as they can be more similar to some of the intermediate representations than to the lower, literal ones.

\section{Discussion and Limitations}\label{sec:disc}
In this Section we would like to discuss the transparency and explainability of our algorithm, its capability of saying ``I do not know'', and its limitations.

Regarding transparency and explainability, it is fundamental to note that, as our algorithm has an internal hierarchical organisation of Sparse Distributed Representations (SDRs), it is possible to recall how our algorithm decided which label corresponds to the input. To that effect, we need the decoder from the Embodiment to transform the internal SDRs into understandable outputs. Thus, we can interpret any decision as a filtering from the Seed Cell, based on its Footprints, and down the hierarchy until the last Footprint that was activated. Then, its representation is the Projection, and the strongest label of that Projection is the selected label.

Regarding the capability of our algorithm to say ``I do not know'', it is easily derived from our threshold setup. If a new input does not surpass the threshold for any Footprint, that is, its similarity with each one of the Footprints in the Nodule is lower than their associated Spatial Attention thresholds, then our algorithm returns a value stating it cannot associate that input to any knowledge it has learned. This is in fact used during training to generate new Footprints. Moreover, that answer is not only an ``I do not know the label'', but it actually means that it does not have a model for such input, so it can not return any Projection of it neither. This is an important and novel feature in an unsupervised learning algorithm. Its importance lies in the fact that saying ``I do not know'' ensures the user understands that the algorithm was not trained to recognise the pattern that was given, instead of falsely providing an answer and hallucinating~\cite{okdg+21,jlfy+23}.

Finally, regarding the limitations of our proposal, its main one is the high memory costs involved compared to other alternatives due to the storage of a huge number of SDRs. We are aware that this limitation can hamper its scalability and applicability over very huge datasets and we are working in ways to diminish it, from developing growth inhibition and death mechanisms for the Footprints and Cells, to improving our Embodiments to generate smaller SDRs.

A secondary but also important limitation is the fact that our proposal is not an optimisation method. This implies that its focus is not to generate the best answer, or to cluster in the best way possible, like other algorithms. Instead, it is focused on building meaningful representations, that are useful to represent the input space, and we expect that this focus will produce, incidentally, a good classifier. This in turn hampers our classification capabilities, and thus our results, but anyway we managed to obtain the good results presented in this paper.

Lastly, it is important to recognise that the first few samples fed to the model are key for the later development of the architecture's structure, producing order and path dependency. Moreover, the Footprint update mechanism produces a stability-plasticity dilemma that can lead to representation drift (internal concept drift). All these problems are derived from our online learning setup. However, on our experiments, none of these problems supposed a limitation, obtaining similar results independently of the samples order. Moreover, if learning goes on for enough time, it's been observed that the impact of the initialization is highly mitigated, and the model converges to offer optimal results. Acknowledging this sensitivity, future work will focus on investigating pre-emptive solutions for the initial seed cell's similarity threshold, as well as introducing reorganization mechanisms to increase the structural stabilization on early steps to maintain the learning agility of the representation model.

\section{Conclusions}\label{sec:conc}
Current well known unsupervised learning methods have a dim capability of extracting cognition-like relationships due to their optimisation oriented setup. The biggest exponent of this field is K-Means, that clusters samples based only on the mathematical distance between them. In this paper we have proposed an alternative, input-agnostic, representation-centric, unsupervised learning algorithm for decision-making that extracts cognition-like relationships between samples through constructive representations.

Our proposal transforms the inputs into SDRs, and then generates an internal representation of those SDRs in order to later recall that representation when asked about the class of an specific input. We tested our proposal against K-Means and IIC for unsupervised classification tasks in four datasets, and show that our proposal is equivalent to them, even although it only process each sample once. Moreover, we have compared it with the state-of-the-art for processing small and incomplete datasets and for identifying cancer types, and overcome all of them. Finally, we have evaluated how well it can discover cognition-like relationships compared to other clustering algorithms, and we have found that it is better than the three main clustering algorithms: K-Means, IIC and K-NN. This is important because it means that our proposal does not only have a different, better behaviour than unsupervised learning algorithms, but also than supervised learning clustering ones.

As future work, we would like to explore how our proposal performs in other datasets and against other unsupervised learning algorithms, and perform an in deep analysis of the relevance of each ``parameter'' of our model. We would also like to develop new Embodiments for different input types, like sound. We would like to explore the extension of our algorithm with other modulators too, like a conditioning modulator that allows us to have a reinforcement learning-like algorithm, or a temporal modulator that allows us to process sequences. We would like to explore different algorithms to compute the similarity function or the spatial attention function too. Finally, we would like to extend our proposal with growth inhibition and death mechanisms for the Footprints and Cells, in order to reduce its memory cost.

\subsubsection{\ackname}
We want to thank Daniel Pinyol for his help building the code of our proposal, and Daniel Pinyol and Pere Mayol for our insightful discussions about the topic.
This work has been supported by the Torres-Quevedo grant PTQ2023-012986 funded by the MCIU/AEI /10.13039/501100011033.

\bibliographystyle{splncs04}
\bibliography{biblio}
\end{document}